\documentclass[final]{cvpr}

\usepackage{amsmath,amsfonts,bm}

\def\eqref#1{equation~\ref{#1}}

\def\1{\bm{1}}

\DeclareMathAlphabet{\mathsfit}{\encodingdefault}{\sfdefault}{m}{sl}
\SetMathAlphabet{\mathsfit}{bold}{\encodingdefault}{\sfdefault}{bx}{n}

\DeclareMathOperator*{\argmax}{arg\,max}

\usepackage{url}
\usepackage[utf8]{inputenc} 
\usepackage[T1]{fontenc}    
\usepackage{booktabs}       
\usepackage{amsfonts}       
\usepackage{nicefrac}       
\usepackage{microtype}      
\usepackage{color}
\usepackage{graphicx}
\usepackage{multirow}
\usepackage{amsmath}
\usepackage{wrapfig}
\usepackage{wrapfig,lipsum,booktabs}
\usepackage[nolist]{acronym}
\usepackage{amsmath}
\usepackage{float}

\usepackage[pagebackref=true,breaklinks=true,colorlinks,bookmarks=false]{hyperref}

\def\cvprPaperID{6646} 
\def\confYear{CVPR 2021}

\begin{document}

\title{Efficient Feature Transformations for Discriminative and Generative Continual Learning}

\author{Vinay Kumar Verma$^{1}$, Kevin J Liang$^{1,2}$, Nikhil Mehta$^{1}$, Piyush Rai$^{3}$, Lawrence Carin$^{1}$\\
$^{1}$Duke University \quad$^{2}$Facebook AI \quad$^{3}$IIT Kanpur\\
{\tt\small vinaykumar.verma@duke.edu}
}

\maketitle

\maketitle

\begin{abstract}
	As neural networks are increasingly being applied to real-world applications, mechanisms to address distributional shift and sequential task learning without forgetting are critical. 
	Methods incorporating network expansion have shown promise by naturally adding model capacity for learning new tasks while simultaneously avoiding catastrophic forgetting.
	However, the growth in the number of additional parameters of many of these types of methods can be computationally expensive at larger scales, at times prohibitively so. Instead, we propose a simple task-specific feature map transformation strategy for continual learning, which we call \emph{Efficient Feature Transformations (EFTs)}. These EFTs provide powerful flexibility for learning new tasks, achieved with minimal parameters added to the base architecture.
	We further propose a feature distance maximization strategy, which significantly improves task prediction in class incremental settings, without needing expensive generative models.
	We demonstrate the efficacy and efficiency of our method with an extensive set of experiments in discriminative (CIFAR-100 and ImageNet-1K) and generative (LSUN, CUB-200, Cats) sequences of tasks.
	Even with low single-digit parameter growth rates, EFTs can outperform many other continual learning methods in a wide range of settings.
\end{abstract}

\vspace{-2mm}
\section{Introduction}
\vspace{-1mm}
While deep learning has led to impressive advances in many fields, neural networks still tend to struggle in sequential learning settings, largely due to catastrophic forgetting~\cite{McCloskey1989, Ratcliff1990}: when the training distribution of a model shifts over time, neural networks overwrite previously learned knowledge if not repeatedly revisited during training. Pragmatically, this typically means that data collection must be completed before training a neural network, which can be problematic in settings like reinforcement learning~\cite{mnih2015human} or the real world~\cite{sculley2015hidden}, which is constantly evolving.
Otherwise, the model must constantly be re-trained as new data arrives.
This limitation significantly hampers building and deploying intelligent systems in changing environments.

\begin{figure*}[!ht]
	\centering
	\includegraphics[scale=0.36]{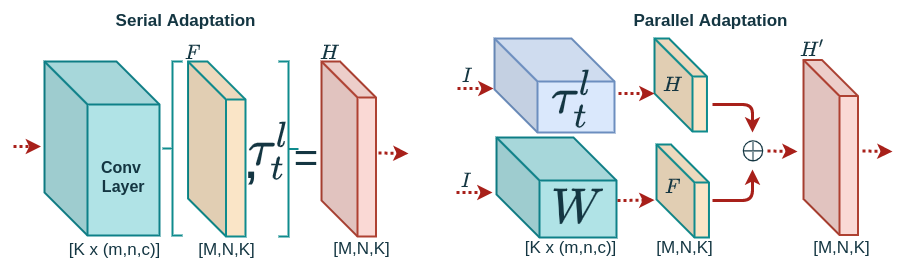}
	\includegraphics[scale=0.36]{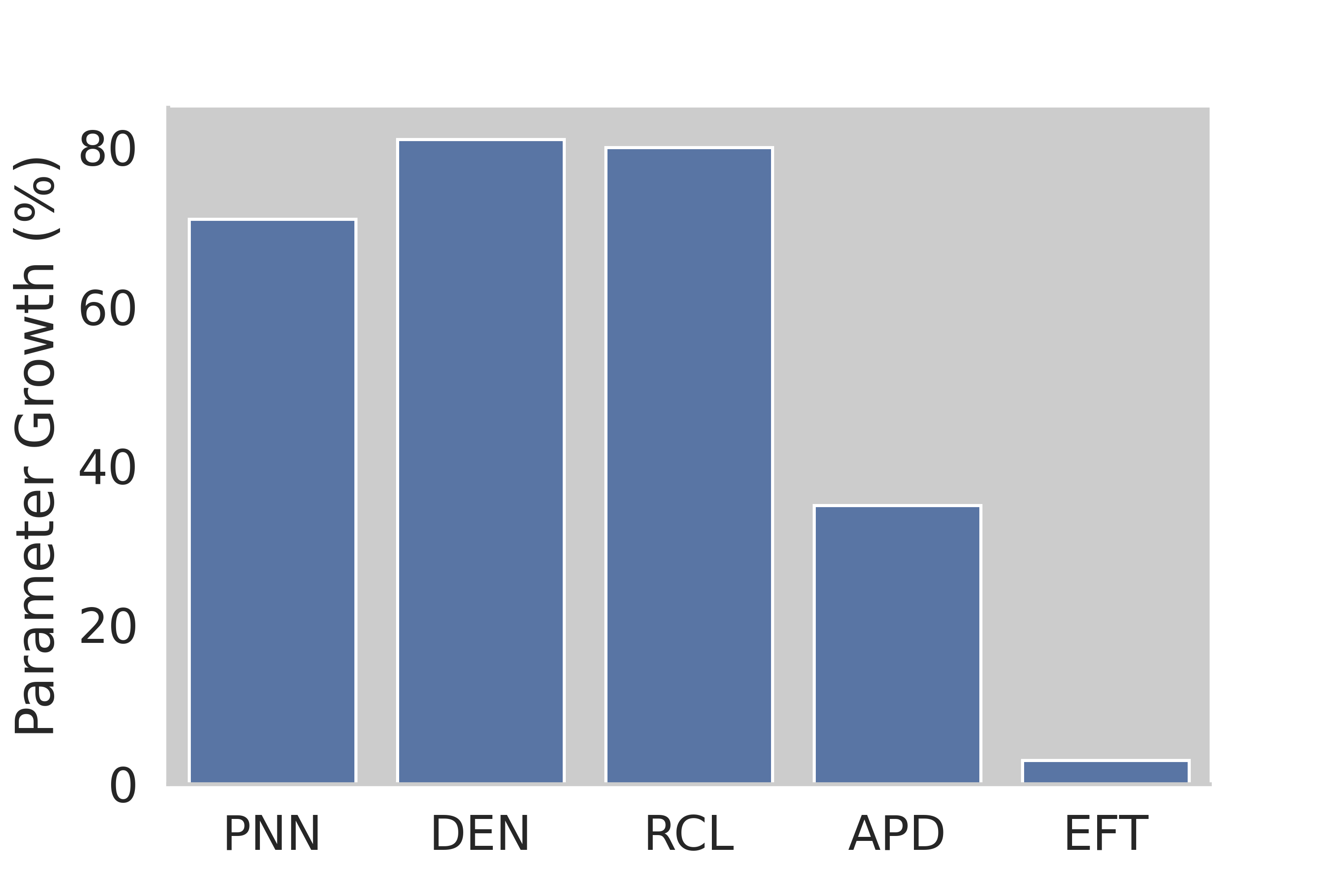}
	\caption{\textbf{Left:} EFT transforms global feature map $F$ to a task-specific feature map $H$ with parameters $\tau_t$. \textbf{Right:} Parameter growth for the 10 task on the LeNet architecture, EFT shows significantly lower growth than other expansion models. }
	\label{fig:hetconv}
	\vspace{-1em}
\end{figure*}

A variety of continual learning methods~\cite{Parisi2019} have been proposed to address this shortcoming.
Regularization-based methods~\cite{Kirkpatrick2017, Zenke2017, Nguyen2018, Aljundi2018, Ritter2018} prevent forgetting by constraining model parameters from drifting too far away from previous solutions, but they can also restrict the model's ability to adapt to new tasks, often resulting in sub-optimal solutions.
Additionally, regularization methods commonly make the assumption that each weight's importance for a task is independent, which may explain why they have difficulty scaling to more complex networks and tasks.
Replay methods~\cite{Lopez2017, Shin2017, Nguyen2018, Rolnick2019} retain knowledge by rehearsing on data saved from previous tasks.
While effective at preventing forgetting, the performance of replay-based approaches is highly dependent on the size and contents of the memory buffer, and in certain strict settings, saving any data at all may not be an option.
The nature of this replay buffer also tends to highly bias the model toward recently learned tasks.
As the number of tasks grows, performance degrades quickly, especially in large-scale settings~\cite{rajasegaran2020itaml, rebuffi2017icarl}.

As an alternative to regularization or replay, expansion methods~\cite{Rusu2016, yoon2017lifelong, yoon2020scalable, mehta2021bayesian} combat forgetting by growing the model with each task.
Expansion alleviates catastrophic forgetting by design, and additional necessary capacity can be easily added to accommodate new knowledge needed for new tasks.
This ability to scale arbitrarily without needing to save any data gives expansion methods the best chance to succeed in large-scale settings.
However, added capacity to model future tasks needs to be carefully balanced against the number of parameters added, especially since the number of tasks the model must learn is often unknown ahead of time. Too inefficient of an approach can easily exceed computational resources even after a moderate number of tasks. Moreover, many previous expansion methods are either computationally inefficient or inflate model size by a significant amount.

To overcome these limitations, we propose a compact, task-specific feature map transformation for large-scale continual learning, which we call \emph{Efficient Feature Transformation (EFT)}.
In particular, we partition the model into \emph{global} parameters ($\theta$) and task-specific 
\emph{local} parameters ($\tau_t$), with the pair ($\theta,\tau_t$) as the optimal parameters for a particular task $t$ (see Figure~\ref{fig:hetconv}). 
In constructing these local transforms, we leverage efficient convolution operations \cite{howard2017mobilenets, singh2019hetconv, xie2017aggregated}, maintaining expressivity, while keeping model growth small.
We also minimize the impact on the global base architecture, allowing us to use pre-existing architectures, which can be critical for achieving strong performance in large-scale settings.
This compact nature of the added transformations also makes EFTs faster to train than comparable methods because we have to update only task-specific parameters.
Finally, we propose a strategy for maximizing feature distance to improve task prediction, a critical component for continual learning methods operating in class incremental settings.

To show the efficacy and efficiency of the proposed approach, we extensively evaluate our model on a variety of datasets and architectures. 
In class incremental and task incremental sequential classification settings, EFTs achieve significant performance gains on CIFAR-100~\cite{Krizhevsky2009} and ImageNet~\cite{Deng2009} with only a minor growth in parameter count and computation. 
We also evaluate our approach for continual generative modeling, demonstrating a $22.7\%$ relative improvement in FID~\cite{heusel2017gans} on the LSUN~\cite{yu15lsun}, CUB-200~\cite{WelinderEtal2010}, and ImageNet~\cite{Deng2009} cat datasets compared to recent state-of-the-art models. 

\vspace{-1mm}
\section{Methods}
\vspace{-1mm}
\subsection{Efficient Feature Transforms}
\vspace{-1mm}
Commonly used modern vision and language architectures can be quite large, sometimes having tens to hundreds of millions of parameters~\cite{He2016, devlin2019bert}.
This can make them incompatible with many previous continual learning methods: if one must add entire parallel columns~\cite{Rusu2016} or regularize the entirety of a model's weights~\cite{Kirkpatrick2017} per task $t$, it can very easily exceed system capacity.
Additionally, many of large-scale networks have been carefully engineered through years of manual architecture search, resulting in structures with specific hyperparameter settings or inductive biases that make them especially effective~\cite{cohen2016inductive}.
Modifications to these design may lead to unintentional degradation of the model's overall effectiveness.
Many previous continual learning methods significantly alter the base network in a way that may be detrimental to overall performance.

With these considerations in mind, we propose partitioning the network into a global base network parameterized by $\theta$ and task-specific transformations $\tau_t$.
During a task $t$, only the task-specific parameters $\tau_t$ are trained; previous local parameters $\tau_{<t}$ and global parameters $\theta$ remain unchanged.
Under this set-up, the global network can be any architecture, preferably an effective one for the problem at hand. 
Of particular note, this means that pre-trained weights can also be employed, if available and appropriate.
Because attempting to transform parameters $\theta$ in its entirety can be expensive, we instead propose task-specific local transformations of the \textit{features} at various layers within the network.
We aim to keep the task-specific transformations minimal.
In particular, using efficient operations, our approach can keep the number of parameters in $\tau_t$ very small without degrading performance. 
To ensure network compatibility, we also ensure that the dimensionality of each transformed feature tensor remains the same as the original, meaning that this operation can be inserted into any architecture without any changes.
We outline here how this can be done for 2D convolutional and fully connected layers.
We focus on these two operations as they comprise the backbone of many deep architectures, but these concepts can be generalized to other types of layers as well.

\vspace{-1mm}
\subsubsection{2D Convolutional Layer}
\vspace{-1mm}
Convolutional neural networks (CNNs)~\cite{Lecun1998} play a major role in many modern computer vision algorithms, with the 2D convolutional layer being one of the primary building blocks.
Let $I$ be the input to a convolutional layer.
In the typical formulation, each convolution layer is composed of $K$ convolutional filters $\mathcal W$; each filter $\mathcal W_k$ in $\mathcal W$ has size $(m \times n \times c)$, with $m$ and $n$ being the filter's spatial dimensions and $c$ the number of channels in $I$.
Each filter $\mathcal W_k$ convolved with $I$ produces an output feature map $F_k \in \mathbb R^{M\times N}$. 
The whole operation can be summarized as
\begin{equation}
	F = \mathcal W * I
\end{equation}
with $*$ being the 2D convolution operator and $F \in \mathbb R^{M\times N \times K}$ being all feature maps $F_k$ stacked into a tensor.
Generally, this operation includes an additive bias, which we omit here for notational convenience.

Rather than adjusting $\mathcal W$ directly, we instead propose appending a small convolutional transformation to change the features $F$.
As this operation needs to be done for each task $t$, we seek to do so efficiently, more so than existing expansion-based methods.
We do this by leveraging groupwise~\cite{krizhevsky2012imagenet, xie2017aggregated} and depthwise~\cite{howard2017mobilenets} convolutions, producing new features for a specific task.
We use two types of convolutional kernels: $\omega^s \in \mathbb R^{3 \times 3 \times a}$ for capturing spatial features within groups of channels and $\omega^d \in \mathbb R^{1 \times 1 \times b}$ for capturing features across channels at every pixel in $F$, where $a$ and $b$ are hyperparameters defining the group size.
For the former case, we perform a groupwise convolution with cardinality $a$ using depth-$a$ convolutions.
In other words, we split convolutional feature maps $F$ into $K/a$ groups, and for each spatially convolve a unique set of $a$ filters $\omega_i^s$ over the group of feature maps.
The resulting $K/a$ groups of $a$ feature maps $H_i^s$, are all concatenated into a single tensor $H^s \in \mathbb R^{M\times N \times K}$.
Importantly, this is the same dimensions as the original feature map $F$, keeping with our goal of keeping the architecture unchanged after inserting this transformation.
A similar operation is done for filters $\omega^d$, but with cardinality $b$ and depth $b$, to constitute feature maps $H^d \in \mathbb R^{M\times N \times K}$.
The construction of both feature maps can be expressed as:
\vspace{-0.5em}
\begin{align}
	\begin{split}
		H^s_{ai:(ai+a-1)} &= [\omega^s_{i, 1} * F_{ai:(ai+a-1)} ~|~ \dots ~|~ \\& \omega^s_{i, a} * F_{ai:(ai+a-1)}], \quad i \in \{0, \dots, \frac{K}{a}-1\}
	\end{split}\\
	\begin{split}
	H^d_{bi:(bi+b-1)} &= [\omega^d_{i, 1} * F_{bi:(bi+b-1)} ~|~ \dots ~|~ \\& \omega^d_{i, b} * F_{bi:(bi+b-1)}], \quad i \in \{0, \dots, \frac{K}{b}-1\}
	\end{split}
\end{align}
where $|$ is the concatenation operation, $H^s_{ai:(ai+a-1)}\in \mathbb R^{M\times N \times a}$, and $H^d_{bi:(bi+b-1)}\in \mathbb R^{M\times N \times b}$.
Generally, $a \ll K$ and $b \ll K$, so this operation is far more parameter efficient than simply using another convolutional layer with $K$ filters.
The latter requires at least as many parameters as the original network itself for each task, the same as learning separate models per task.

We combine the features from the spatial and depth convolutions $H^s$ and $H^d$ additively:
\vspace{-0.5em}
\begin{equation}\label{eq:add_H}
	H = H^s + \gamma H^d
\end{equation}
where $\gamma \in \{0,1\}$ is an indicator indicating if the pointwise convolutions $\omega^d$ are employed.
Note, if $\gamma=0$, we do not perform convolutions with $\omega^d$; this sacrifices some expressivity, but further reduces the number of added parameters per task.
For example, the extreme case of $a=1$ and $\gamma=0$ results in a small $0.17\%$ increase in parameters per task in ResNet-18 while still achieving state-of-the-art performance on the ImageNet-1K/10 continual learning task.
\begin{figure}[t]
	\centering
	\includegraphics[scale=0.25]{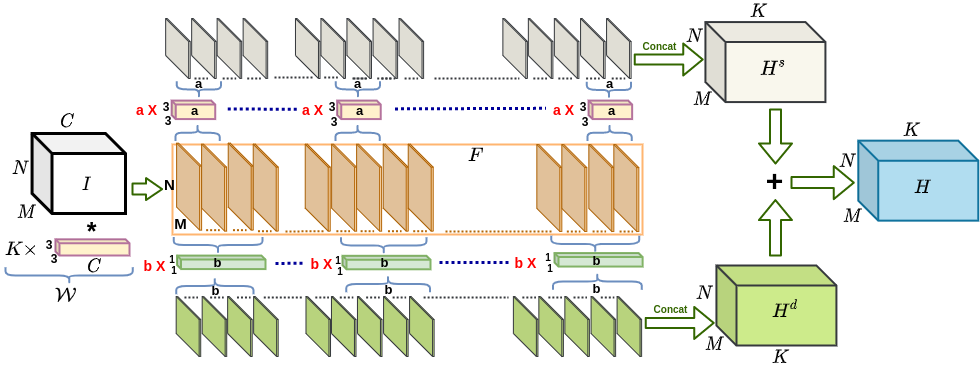}
	\vspace{-1em}
	\caption{Illustration of the proposed parameter-efficient transformation of convolutional features.  
	}
	\label{fig:my_label}
	\vspace{-.5em}
\end{figure}
\vspace{-1mm}
\subsubsection{Fully Connected Layer}
\vspace{-1mm}
Fully connected layers are common in many architectures, typically to project from one dimensionality to another.
In convolutional neural networks, they are frequently used to project to the number of output classes to produce the final prediction.
A fully connected layer is implemented as a matrix multiply between a weight matrix $W$ and an input vector $v$, producing an output feature vector $f = Wv$; the bias vector is again omitted for convenience.

As with the convolutional layer, where we transform the convolutional features $F$ to a task-specific feature $H$ with additional convolutional operations, we also transform output vector $f$ with another fully connected layer (parameterized by $E$) to a task-specific feature vector $h$.
Like in the convolutional case, we must put restrictions on the form of $E$ to prevent this operation from being overly costly.
In particular, we constrain $E$ to be diagonal, which significantly reduces the number of parameters.
This operation can be expressed as $h = Ef$.
In practice, since $E$ is diagonal, this operation can be implemented as a Hadamard product.

\vspace{-1mm}
\subsubsection{Parallel Adaption}\label{sec:parallel}
\vspace{-1mm}
We have thus far introduced our approach as an efficient, sequential feature transformation, adapting the features to each task \textit{after} its computation in the base network.
An alternative parameterization is to compute these feature calibrations in parallel (Figure~\ref{fig:hetconv}, center), making the transformation additive rather than compositional:
\begin{align}
H'&= (\mathcal W * I)\oplus (\tau_t^l*I)\\
&=F\oplus H
\end{align}  
where $\oplus$ is element-wise addition and $\tau_t^l$ is the task specific parameter at layer $l$. 
Empirically we find that both sequential and parallel models achieve similar performance. Unless otherwise mentioned, we report our results with sequential transformations. 
A comparison of the two types is shown in the supplementary material. 

\vspace{-1mm}
\subsection{Task Prediction}
\label{sec:task_pred}
\vspace{-1mm}
Knowing which task $t$ an inference-time input corresponds to is a common implicit assumption in continual learning, but in some settings this information is unavailable, necessitating predicting the task $t$ alongside the class; such a setting has been referred to as class incremental learning (CIL)~\cite{hsu2019reevaluating}.
Predicting the task ID can be challenging, especially when data from previous tasks is not saved.
We choose to predict $t$ by selecting the task-specific set of parameters $\tau_t$ that produces the maximum confidence prediction for the input.
However, a naive, direct measurement of the post-softmax prediction entropy often performs poorly, as the typical cross-entropy training objective tends to produce high confidence in deterministic neural networks, even for out-of-distribution (OoD) samples~\cite{guo2017calibration}.
To remedy this, we propose feature distance maximization, a simple regularization to increase task prediction ability.

\begin{table*}[!th]
	\small
	\centering
	\addtolength{\tabcolsep}{0.9pt}
	\caption{\small Average accuracy on CIFAR-100 in class incremental learning setting when trained on 10 tasks sequentially.}
	\begin{tabular}{l l  l l l l l l l l l l}
		\toprule
		{Dataset / \#Tasks} &{Methods} & {1} & {2} & 3 & 4 & 5 & 6 & 7 & 8 & 9 & Final \\
		\midrule
		\multirow{9}{*}{CIFAR-100/10} & 
		{Finetune} &  88.5& 47.1& 32.1& 24.9& 20.3& 17.5& 15.4& 13.5& 12.5& 11.4\\
		& FixedRep &  88.5& 45.9& 30.1& 22.4& 17.7& 15.2& 12.3& 11.1& 9.8& 8.8\\
		&LwF~\cite{li2017learning} &  88.5& 70.1& 54.8& 45.7& 39.4& 36.3& 31.4& 28.9& 25.5& 23.9\\
		&EWC~\cite{Kirkpatrick2017} &  88.5 & 52.4& 48.6& 38.4& 31.1& 26.4& 21.6& 19.9& 18.8& 16.4\\
		&EWC+SDC~\cite{yu2020semantic} &88.5&78.8&75.8&73.1&71.5&60.7& 53.9&43.5&29.5&19.3\\
		&SI~\cite{Zenke2017}  &  88.5& 52.9& 40.7& 33.6& 31.8& 29.4& 27.5& 25.6& 24.7& 23.3\\
		&MAS~\cite{Aljundi2018} &  88.5& 42.1& 36.4& 35.1& 32.5& 25.7& 21.0& 19.2& 17.7& 15.4\\
		&RWalk~\cite{chaudhry2018riemannian}  &  88.5& 55.1& 40.7& 32.1& 29.2& 25.8& 23.0& 20.7& 19.5&17.9\\
		&DMC~\cite{zhang2020class}  & 88.5& 76.3& 67.5& 62.4& 57.3& 52.7& 48.7& 43.9& 40.1&36.2\\
		\cmidrule{1-12}
		&EFT-$a_4b_{0}$ (+1.7\%) &  90.2&75.7&68.6&62.6&58.1&53.9&51.6&48.7&46.7&\textbf{44.4}\\
		&EFT-$a_4b_8$ (+2.0\%) & 90.1 &75.5&68.9&63.6&58.7&54.6&52.4&49.8&47.3&\textbf{45.1}\\
		&EFT-$a_8b_{16}$ (+3.9\%) &  90.2 & 76.2&70.1&63.1& 57.9&53.6&52.1&49.6& 47.6&\textbf{45.5}\\
		\bottomrule
	\end{tabular}
	\label{tab:ImageNetCIFAR}
\end{table*}

Without intervention, the pre-output layer feature distributions learned by the parameters $\tau_t$ for each task $t$ may show overlap, which may hinder attempts to discriminate task features.
To mitigate this, we add a regularizer to create a margin between the features produced by each task's $\tau_t$ for a given task's data:
\begin{equation}
\mathcal{L_M}=\sum_{i=1}^{t-1}\max(\Delta-\mathrm{KL}(P_t||Q_i),0)
\label{eq:5}
\end{equation}  
where $P_t=\mathcal{N}(\mu_t,\Sigma_t)$ and $Q_i=\mathcal{N}(\mu_i,\Sigma_i)$ are the distributions of the current task $t$ and earlier task $i<t$.
This regularizer helps the model learn representations for $\tau_t$ such that the current task data ($\mathcal{D}_t$) has at least $\Delta$ separability in the feature space from the features encoded by previous tasks' parameters $\tau_{<t}$. 

\vspace{-1mm}
\subsection{Summary} 
\vspace{-1mm}
The joint loss for Efficient Feature Transforms (EFTs) is defined as:
\begin{equation}
	\small
	\mathcal{L}_{\mathrm{EFT}}=\mathcal{L}(\hat{y},y)+\lambda\mathcal{L_M}
	\label{eq:6}
\end{equation}    
where $\mathcal{L}(\hat{y},y)$ is the standard cross-entropy loss between the predicted class $\hat{y}$ and ground truth $y$, and $\lambda$ is a weighting hyperparameter. At test time, for a given input $x$, we measure the output entropy with each $\tau_t$ and predict the task ID as $\argmax_t(x|\theta,\tau_t)$. Once we have the task ID, we can choose the task-specific parameter $\tau_t$ to predict a class.

In summary, we recognize that if one adjusts the original model parameters in a given layer (what we call ``global'' model parameters), this leads to an adjustment in the features output from the layer. However, the number of layer-dependent global parameters may be massive, and adjustment of them may cause loss of information accrued from prior datasets and tasks. Since layer-wise adjustment of global parameters simply adjusts the output feature map, we leave the global layer-dependent features unchanged, and introduce a new set of lightweight task-specific parameters, that directly refine the feature map itself. These task-specific parameters may adjust to new tasks, while maintaining the knowledge represented in well-training global model parameters. We have developed methods to apply this concept to convolutional and fully-connected layers. Additionally, we make the model more amenable to task ID prediction by maximizing a margin between task-specific feature distributions.

\vspace{-1mm}
\section{Related Work}
\vspace{-1mm}
The continual learning literature is vast~\cite{de2019continual,Parisi2019}. 
Broadly, continual learning (CL) methods can be grouped into three categories, by strategy: replay, regularization, and expansion. 
We focus here primarily on expansion-based methods, as they are the most closely related to our work.

A number of previous works~\cite{zhang2019side,yoon2017lifelong,Rusu2016,xu2018reinforced,gao2020efficient,serra2018overcoming,mallya2018packnet,masana2020ternary,von2019continual,yoon2020scalable} have proposed methods for expanding a neural network to learn a sequence of tasks. 
For example, Progressive Neural Networks (PNNs)~\cite{Rusu2016} adds a new neural network column for each new task; weights for previous tasks are frozen in place, and lateral connections are added for forward knowledge transfer.
Side-tuning~\cite{zhang2019side} takes an approach similar to but simpler than PNNs: they propose learning small task-specific networks whose outputs are fused to the larger base network. Dynamically Expandable Network (DEN)~\cite{yoon2017lifelong} optimizes three sub-problems of selective training, network expansion, and network duplication, while Reinforced Continual Learning RCL~\cite{xu2018reinforced} using reinforcement learning to determine the growth of the architecture.
Alternatively, some recent works leverage Bayesian nonparametrics to let the data dictate expansion~\cite{mehta2021bayesian, Kumar2019, Lee2020}, but the benchmarks considered in Bayesian methods have been limited to MNIST and CIFAR-10. 

An iterative training and pruning strategy~\cite{mallya2018packnet, mallya2018piggyback, hung2019compacting} has also been proposed for expansion-based continual learning, with the pruning incorporated to reduce model growth; however, PackNet~\cite{mallya2018packnet} still requires saving masks to recover networks of previous models, which can take up storage space as the number of tasks grows.
Other mask-based approaches have proven popular in recent years.
HAN~\cite{serra2018overcoming} proposes hard attention masks for the each task.
TFM~\cite{masana2020ternary} applies ternary masks to the feature maps, which results in less memory per mask, as the feature maps are often smaller in size than the number of weights in the model.
Additive Parameter Decomposition (APD)~\cite{yoon2020scalable} uses masks to decompose the model parameters into task-shared and task-specific parameters; however, the significant changes made to architecture makes it harder to scale and precludes the use of pre-trained weights.
SupSup~\cite{wortsman2020supermasks} proposes supermasks~\cite{zhou2019deconstructing} for each task, storing the supermask in a fixed size Hopfield network network~\cite{hopfield1982neural}. 
Masking approaches work well for task-incremental settings, but the task prediction required for class incremental learning to select the appropriate mask can be challenging and costly; replay is sometimes relied upon~\cite{mallya2018packnet}. By contrast, our proposed method shows promising results in both scenarios, without relying on replay.

\begin{figure*}[!h]
	\centering
	\includegraphics[scale=0.4]{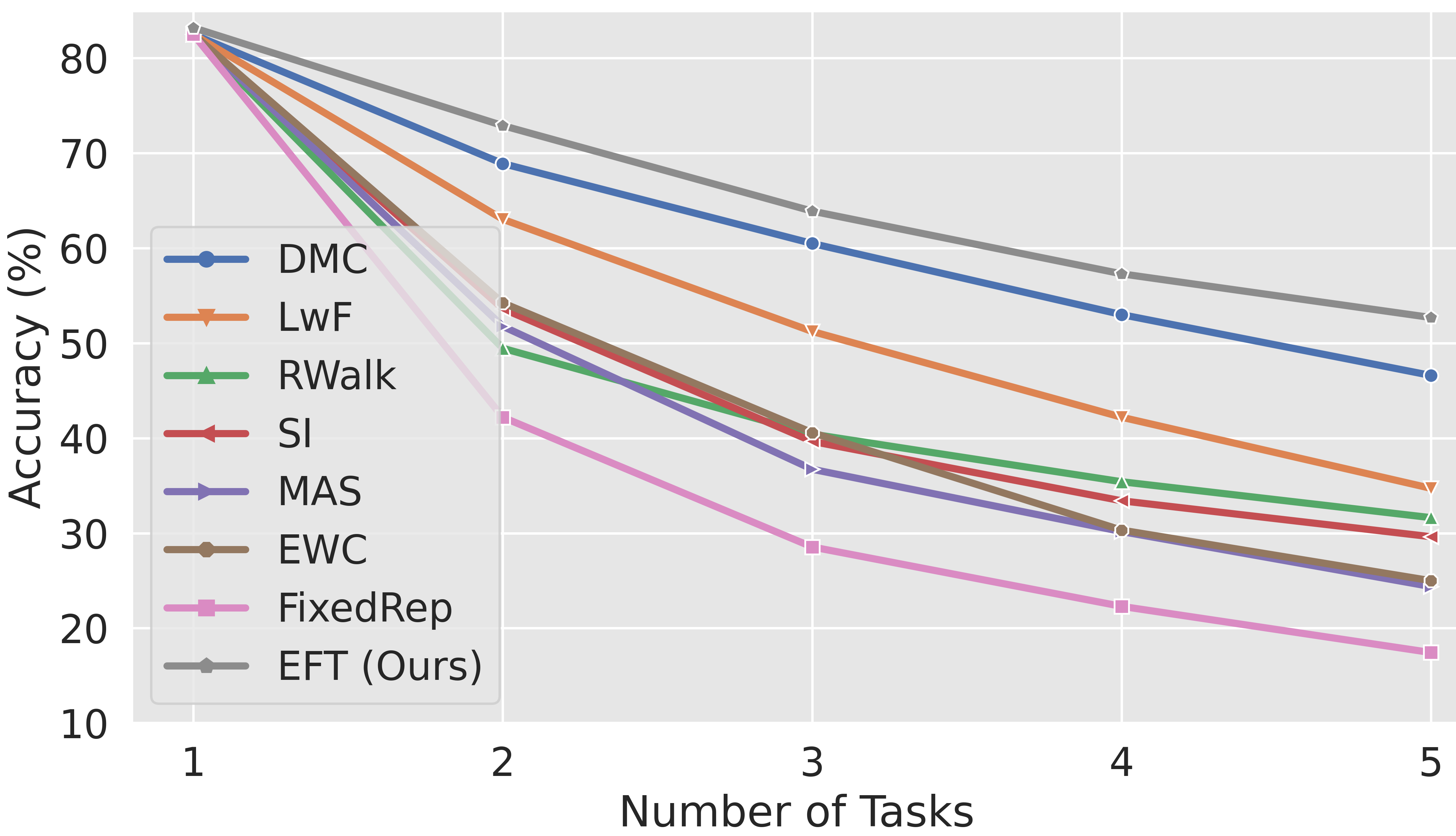} 
	\includegraphics[scale=0.4]{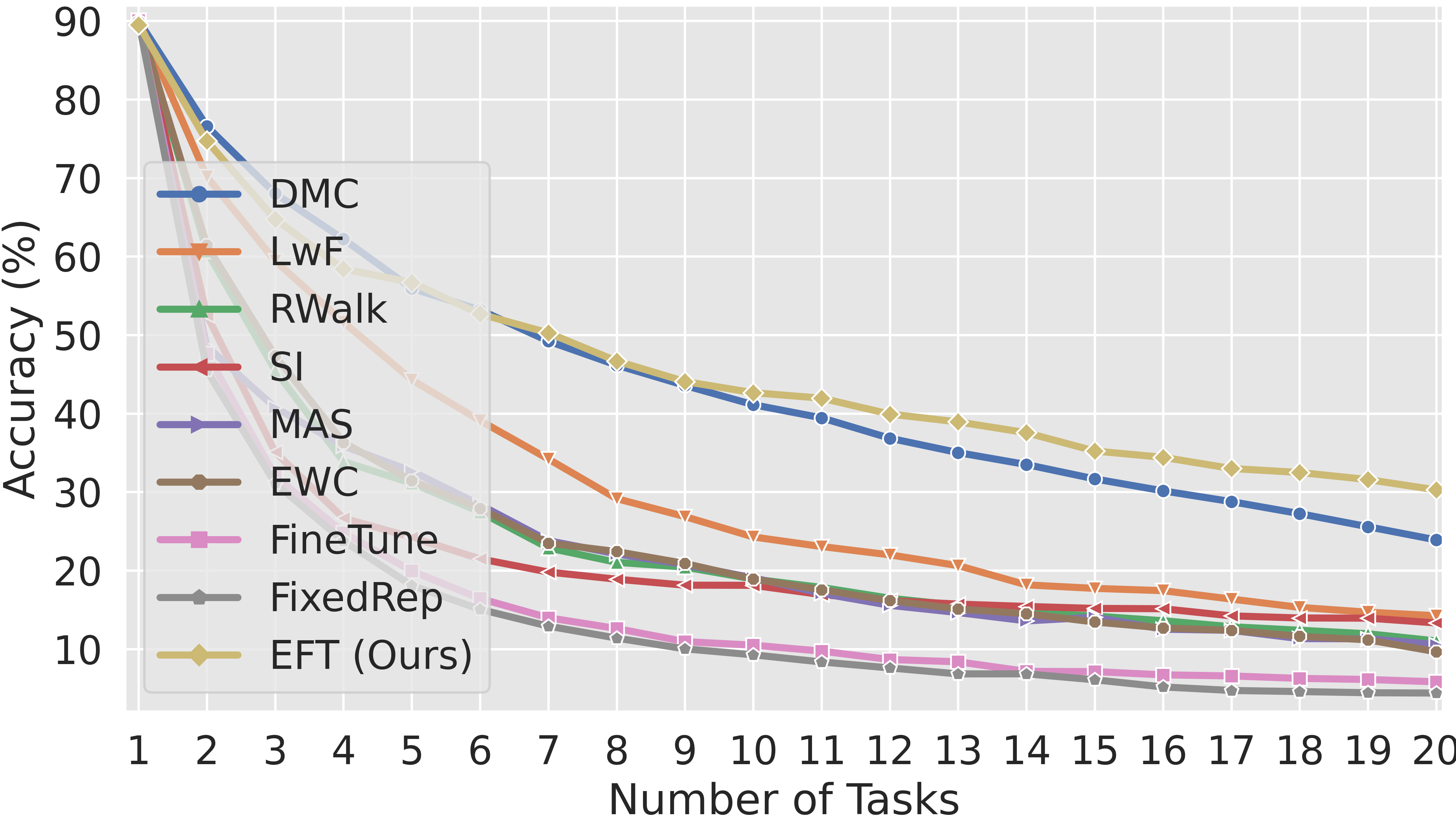}
	\caption{CIFAR-100/5 and CIFAR-100/20 result in the class incremental learning setup.}
	\label{fig:c100_20}
\end{figure*}
\begin{table*}[!th]
	\small
	\centering
	\addtolength{\tabcolsep}{0.5pt}
	\caption{\small Comparison with the state-of-the-art. ImageNet-1K/10 on AlexNet and Tiny-ImageNet-200/10 on VGG-16 from scratch. Accuracy of each task after learning all tasks.}
	\begin{tabular}{l l  l l l l l l l l l l l}
		\toprule
		{Dataset / \#Tasks} &{Methods} & {1} & {2} & 3 & 4 & 5 & 6 & 7 & 8 & 9 & 10 & Avg \\
		\midrule
		\multirow{9}{*}{Tiny ImageNet/10} & 
		Finetuning & 38.1  & 36.0  & 43.2  & 44.1  & 45.5  & 54.5 & 50.3  & 50.5  & 51.0 & 61.2  & 47.4 \\
		& Freezing & 51.7  & 36.4  & 39.5 & 41.7  & 42.9  & 46.2  & 45.7  & 41.1  & 41.2  & 40.9  & 42.7 \\
		&LfL~\cite{jung2016less} & 32.4  & 35.4  & 43.4  & 44.1 & 45.0  & 55.9  & 49.4  & 51.1  & 58.6  & 61.4  & 47.7 \\
		&LwF~\cite{li2017learning} & 45.1  & 45.5  & 53.5  & 57.6  & 56.2  & 65.7  & 63.5  & 58.4  & 59.6  & 58.5  & 56.4 \\
		&IMM-mode~\cite{lee2017overcoming} & 50.6  & 38.5  & 44.7  & 49.2  & 47.5  & 51.9  & 53.7  & 47.7  & 50.0  & 48.7  & 48.3 \\
		(VGG-16)&EWC~\cite{Kirkpatrick2017} & 33.9  & 35.4  & 43.6  & 46.7  & 49.5  & 52.5  & 47.8  & 50.2  & 56.6  & 61.4  & 47.8 \\
		&HAT~\cite{serra2018overcoming} & 46.8  & 49.1  & 55.8  & 58.0  & 53.7  & 61.0  & 58.7  & 54.0  & 54.6  & 50.3  & 54.2 \\
		&PackNet~\cite{mallya2018packnet} & 52.5  & 49.7  & 56.5  & 59.8  & 55.0  & 64.7  & 61.7  & 55.9  & 55.2  & 52.5  & {56.4} \\
		&TFM~\cite{masana2020ternary}  & 48.2  & 47.7  & 56.7  & 58.2  & 54.8  & 62.2  & 61.5  & 57.3  & 58.5  & 54.8  & 56.0 \\
		\cmidrule{2-13}
		&EFT-$a_8b_{16}$ (3.1\%)  & 67.2& 62.5& 69.4& 62.6& 68.3& 69.6& 59.0& 67.8& 71.5& 70.1 &\textbf{66.8} \\
		\midrule
		
		\multirow{5}{*}{ImageNet-1K/10} & 
		Finetuning & 25.8  & 32.2  & 31.4  & 37.8  & 39.1  & 43.7  & 46.0  & 50.0  & 53.4  & 63.7  & 42.3 \\
		&Freezing & 68.8  & 53.5  & 52.0  & 51.2  & 51.3 & 53.9  & 52.2  & 53.9  & 51.7  & 51.2  & 54.0 \\
		&LwF~\cite{li2017learning} & 27.6  & 37.2  & 42.0  & 44.4  & 50.5 & 56.6  & 57.9  & 61.2  & 62.0  & 62.7  & 50.2 \\
		(AlexNet)&IMM-mode~\cite{lee2017overcoming} & 68.5  & 53.6  & 52.1  & 51.7  & 52.5  & 55.5  & 54.7  & 53.5  & 54.2  & 51.8  & 54.8 \\
		&EWC~\cite{Kirkpatrick2017} & 21.8  & 26.5  & 29.5  & 32.9  & 35.6  & 40.4  & 40.0  & 44.7  & 47.8  & 61.1  & 38.0 \\
		&PackNet~\cite{mallya2018packnet} & 67.5 & 65.8  & 62.2  & 58.4  & 58.6  & 58.7  & 56.0  & 56.5  & 54.1  & 53.6  & 59.1 \\
		\cmidrule{2-13}
		&EFT-$a_{16}b_{64}$ (0.6\%) & 69.0&63.2&60.1&62.5&53.6&57.2&55.1&52.8&55.7&62.5& \textbf{59.4} \\
		\bottomrule	
		
	\end{tabular}
	\label{tab:ImageNet_tiny}
\end{table*}

\vspace{-2mm}
\section{Experiments}
\vspace{-2mm}
We demonstrate our approach's performance for multiple base architectures (ResNet~\cite{He2016}, VGG~\cite{simonyan2014very}, AlexNet~\cite{krizhevsky2012imagenet}) and datasets (ImageNet-1K~\cite{Deng2009}, CIFAR-100~\cite{Krizhevsky2009}, Tiny-ImageNet) in class and task incremental learning scenarios.\footnote{\url{https://github.com/vkverma01/EFT}} 
We use the nomenclature \texttt{[DATASET]}-$\mathcal C$/$T$ throughout this section to denote a continual learning set-up with $\mathcal C$ total classes evenly divided into $T$ tasks, meaning each task involves learning $\frac{\mathcal C}{T}$ new classes.
We also explore generative continual learning on the StackGAN-v2~\cite{zhang2018stackgan} architecture for the four sequential task (cats, birds, church and tower), and run several ablation studies. In all scenarios, we achieve a significant improvement compared to the base model.

\vspace{-1mm}
\subsection{CIFAR-100}
\vspace{-1mm}
CIFAR-100~\cite{Krizhevsky2009} is commonly among the most challenging datasets considered by many previous continual learning methods. 
We break down the 100 classes into three different class sequence splits: CIFAR-100/5, CIFAR-100/10 and CIFAR-100/20.
With more classes per tasks, CIFAR-100/5 requires the model to learn a harder problem for each task, while CIFAR-100/20 increases the length of the task sequence, testing a continual learning method's retention. 
We run our CIFAR experiments in the class incremental setting, which means task information is unknown at test time, thus requiring task prediction.
In our CIFAR experiments, we utilize the ResNet-18~\cite{He2016} architecture for CIFAR datasets as our base architecture. 
We report the average top-1 accuracy of all previously seen tasks up to $t$, for each $t$, averaged over 5 random permutations of task order.
We compare EFT with several other popular continual learning methods for CIFAR-100/10 in Table~\ref{tab:ImageNetCIFAR}, and plot performance over time (tasks seen) for CIFAR-100/5 and CIFAR-100/20 in Figure~\ref{fig:c100_20}.

If simply training on each task dataset in sequence, finetuning the model on each incoming dataset without any continual learning measures, we observe severe catastrophic forgetting.
Because of the diversity and complexity of the classes in each task, previous tasks are forgotten almost entirely in order to specialize the model for the current task.
While other methods show improvements over the finetuning baseline, all still show major signs of forgetting.
We show results for EFTs with different convolutional group sizes $\{a,b\}$, using the notation EFT-$a_{\alpha}b_{\beta}$ to indicate EFT with $a=\alpha$ and $b=\beta$.
For CIFAR-100/10, EFT-$a_4b_0$, EFT-$a_4b_{8}$, and EFT-$a_8b_{16}$ result in small $1.7\%$, $2.0\%$, and $3.9\%$ increases in parameters per task respectively, leading to 8.2\%, 8.9\% and 9.3\% absolute gain in final average accuracy in comparison to recent regularization or expansion-based models.  We observe a similar pattern for CIFAR-100/20 and CIFAR-100/5: we see consistently better results throughout the task sequence when there is a larger number of tasks (CIFAR-100/20) and when individual tasks are harder (CIFAR-100/5). 
More details about the experimental setup and hyperparameters can be found in the supplementary material, as well as
additional comparisons with SupSup~\cite{wortsman2020supermasks} and CCLL~\cite{singh2020calibrating}.

\vspace{-1mm}
\subsection{ImageNet}
\vspace{-1mm}
The ImageNet-1K~\cite{Deng2009} classification dataset contains 1000 classes based on the WordNet~\cite{miller1998wordnet} hierarchy. Tiny-ImageNet is a smaller subset of the ImageNet dataset, containing 200 classes downsampled to $64\times 64\times 3$ resolution. 
For many years, ImageNet-1K served as a measuring stick for deep computer vision progress, and it still remains a very difficult and rarely tested setting for continual learning.
Our ImageNet experiments are conducted in the task-incremental learning scenario, assuming the task-id at test time to choose $\tau_t$ for that task. 
We use VGG-16~\cite{simonyan2014very} and AlexNet~\cite{krizhevsky2012imagenet} CNN architectures for Tiny-ImageNet-200/10 and ImageNet-1K/10, respectively. The results are shown in the Table~\ref{tab:ImageNet_tiny}. We report the average top-1 accuracy across all encountered tasks for three random task orders. 
Compared with the baselines, we see significant performance improvements with EFTs. 
Notably, with VGG-16 on Tiny-ImageNet-200/10, we observe a $10.4\%$ absolute gain on the 10-task sequence. Please refer to the supplementary material for more details about the model architecture and hyperparameter settings.

\begin{table}[!t]
	\centering
	\addtolength{\tabcolsep}{-3.5pt}
	\caption{Final test accuracies after heterogeneous dataset sequence with VGG16. $\downarrow$ follow the SVHN $\rightarrow$ CIFAR-10 $\rightarrow$ CIFAR-100 task order, while $\uparrow$ corresponds to CIFAR-100 $\rightarrow$ CIFAR-10 $\rightarrow$ SVHN.}
	\label{tab:hetero}
	\scalebox{0.8}{
	\begin{tabular}{c|c c|c c|c c|c c|c c}
		\toprule
		{} &\multicolumn{2}{c|}{L2T} & \multicolumn{2}{c|}{PB~\cite{mallya2018piggyback}} & \multicolumn{2}{c|}{PNN~\cite{Rusu2016}} & \multicolumn{2}{c|}{APD~\cite{yoon2020scalable}} &\multicolumn{2}{c}{EFT-$a_{16}b_{16}$} \\
		\toprule
		Task order & $\downarrow$&$\uparrow$ &$\downarrow$&$\uparrow$ &$\downarrow$&$\uparrow$ &$\downarrow$&$\uparrow$&$\downarrow$&$\uparrow$\\
		\toprule
		SVHN & 10.7& 88.4& 96.8& 96.4& 96.8& 96.2& 96.8& 96.8 &96.8&95.5\\
		CIFAR-10 &41.4& 35.8& 83.6& 90.8& 85.8& 87.7& 90.1& 91.0&89.2&90.4\\
		CIFAR-100 & 29.6& 12.2& 41.2& 67.2& 41.6& 67.2& 61.1& 67.2&64.6&71.5\\
		\midrule
		Avg &  27.2& 45.5& 73.9& 84.8& 74.7& 83.7& 83.0& 85.0 &\textbf{83.5}&\textbf{85.8}\\
		\bottomrule
	\end{tabular}
	}
	\vspace{-2mm}
\end{table}

\begin{table}
	\begin{minipage}{0.46\columnwidth}
	\centering
	\addtolength{\tabcolsep}{0pt}
	\caption{Expansion cost and CIFAR-100/10 average accuracy with the convolutional architecture in~\cite{yoon2020scalable}. 
	}
	\scalebox{0.9}{
		\begin{tabular}{ l l l}
			\toprule
			{Methods} & Cost & Acc. \\
			\midrule
			PNN~\cite{Rusu2016} &1.71x& 54.90\\
			DEN~\cite{yoon2017lifelong} &1.81x& 57.38\\
			RCL~\cite{xu2018reinforced}& 1.80x& 55.26\\
			APD~\cite{yoon2020scalable} &1.35x& 60.74\\
			\midrule
			EFT &  1.04x& \textbf{64.17}\\
			\bottomrule
	\end{tabular}}
	\vspace{-2mm}
	\label{tab:capacity}
\end{minipage}
\hfil \hspace{1em}
\begin{minipage}{0.46\columnwidth}
	\centering
	\addtolength{\tabcolsep}{-1pt}
	\caption{Feature distance maximization ablation for CIFAR-100/10 with ResNet-18: CIL and task prediction (TP) accuracies after 10 tasks.
	}
	\scalebox{0.9}{
		\begin{tabular}{l c c}
			\toprule
			Models&  CIL & TP\\
			\toprule
			$a_8b_{16}$& 44.4&44.9\\
			$a_8b_{16}+\mathcal{L_M}$&45.5&46.1\\
			$a_4b_{8}$&44.0&44.3\\
			$a_4b_{8}+\mathcal{L_M}$&45.1&45.7 \\
			\bottomrule
	\end{tabular}}
	\vspace{-2mm}
	\label{tab:ablation}
\end{minipage}
\end{table}

\subsection{Heterogeneous Datasets}
\vspace{-1mm}
While the classes of ImageNet-1K and CIFAR-100 cover a diverse set of categories, tasks drawn solely from CIFAR-100/10 or ImageNet-1K/10 ultimately share many similarities: for example, they mostly consist of natural images, and the total number of classes per task are the same.
It can also be of interest to evaluate if continual learning methods can adapt to shifts in domain or label space cardinality.
To test this, we evaluate our method on Street View House Numbers (SVHN)~\cite{netzer2011reading}, CIFAR-10, and CIFAR-100, using VGG-16~\cite{simonyan2014very}.
Between SVHN and CIFAR-10, the model must adapt from digits to animals and vehicles, and between CIFAR-10 and CIFAR-100, the model must adapt between a 10-class and a 100-class problem.

Results training on this sequence of tasks are shown in Table~\ref{tab:hetero}, for both directions.
Once again, compared to the baselines, we observe that our approach does a significantly better job retaining performance on old tasks while also maintaining plasticity for learning new ones.
Additionally, previous work~\cite{yoon2020scalable} has highlighted the potential importance of task-order robustness in continual learning, which can play a role in model fairness.
We observe that our method has low order sensitivity as well, with similar accuracies on each dataset regardless of the order the datasets are learned.

\begin{figure}
	\vspace{-0em}
	\begin{center}
		\includegraphics[width=0.47\textwidth]{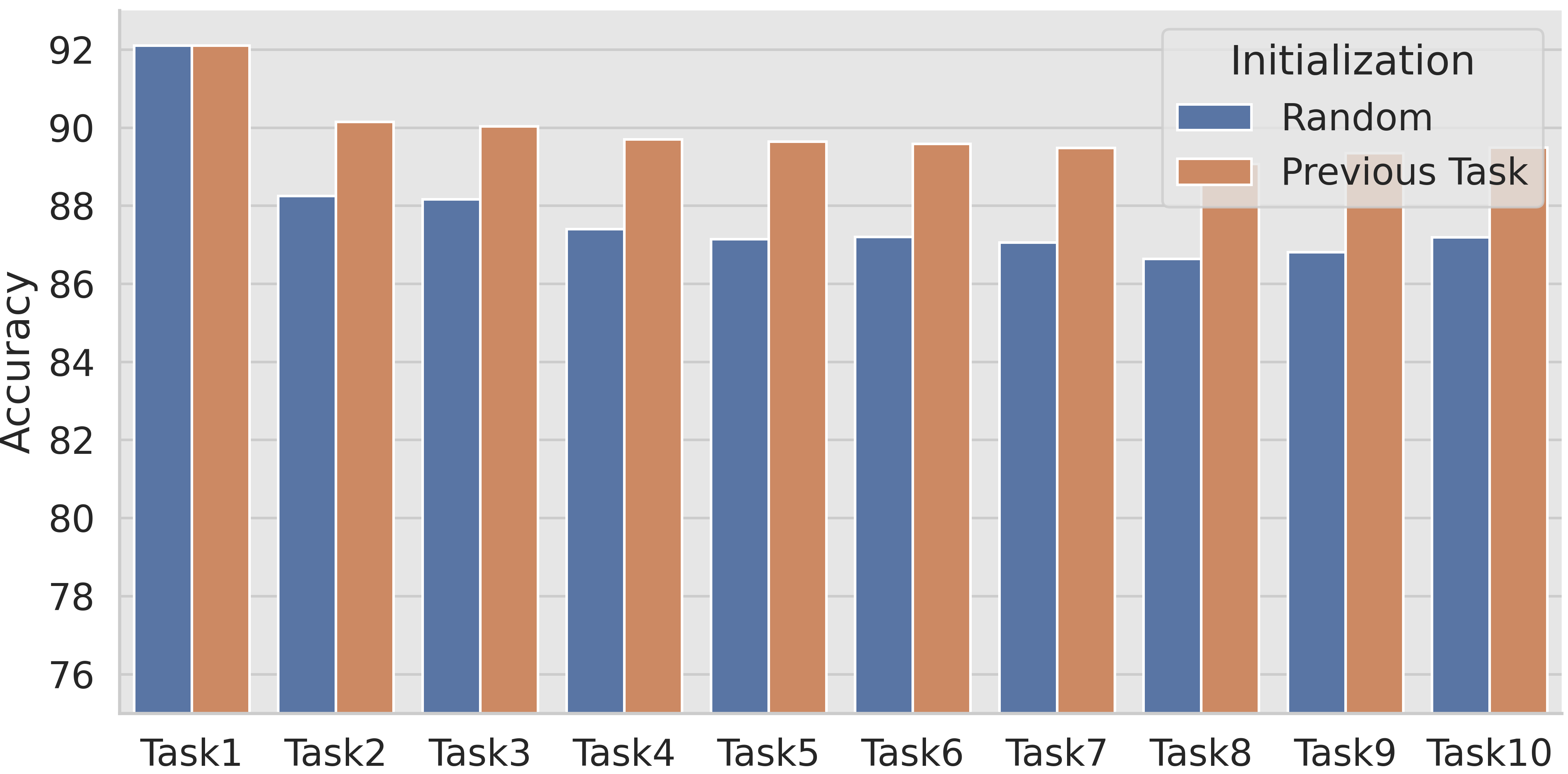}
	\end{center}
	\vspace{-1em}
	\caption{Accuracy after task $t$, with and without forward transfer. }
	\label{fig:forward_transfer}
\end{figure}

\subsection{Forward Transfer}
Positive forward knowledge transfer is an essential ability in continual learning systems. Successful transfer leverages previously learned knowledge, reducing the amount of new data and training time necessary to learn future tasks.
EFTs achieve forward transfer by initializing the task-specific parameters $\tau_t$ with the previous task local parameters $\tau_{t-1}$. We evaluate the forward transfer of EFTs by comparing this forward transfer approach with a random initialization strategy for $\tau_t$, plotting the performance of the model on CIFAR-100/10 using ResNet-18 ($a_8b_{16}$) in Figure~\ref{fig:forward_transfer}.
Empirically, we observe that initializing $\tau_t$ from $\tau_{t-1}$ results in consistently better performance, verifying that our approach results in positive forward transfer.

\begin{table}
	\centering
	\addtolength{\tabcolsep}{1pt}
	\caption{Computation for CIFAR-100/10 with ResNet-18. 
	}
	\label{tab:flops}
	\scalebox{0.9}{
		\begin{tabular}{l l l c}
			\toprule
			Models& {Paramater ($\uparrow )$} & FLOPs ($\uparrow$) & Accuracy \\
			\toprule
			$Base$& 11.17M & 1.11G &--\\
			$a_8b_{16}$& 11.60M (3.87\%)& 1.21G (8.78\%) & 45.5\\
			$a_4b_{8}$& 11.39M (1.97\%)& 1.16G (4.44\%) & 45.1\\
			$a_4b_{0}$& 11.35M (1.59\%) & 1.15G (3.60\%) &44.4\\
			$a_0b_{64}$& 11.48M (2.79\%) & 1.18G (6.36\%) &43.2\\
			$a_0b_{32}$& 11.33M (1.42\%) & 1.15G (3.20\%) &42.3\\
			\bottomrule
	\end{tabular}}
	\vspace{-2mm}
\end{table}

\begin{table*}[!th]
	\centering
	\small
	\addtolength{\tabcolsep}{-1.5pt}
	\caption{GAN FIDs evaluated after training on each dataset, sequentially}
	\label{tab:gan}
	\begin{tabular}{l | c c c c | c c c c | c c c c | c c c c | c}
		\toprule
		& \multicolumn{4}{c|}{Cats} & \multicolumn{4}{c|}{Birds} & \multicolumn{4}{c|}{Churches} & \multicolumn{4}{c|}{Towers} & Final \\
		Task $t$ & 1 & 2 & 3 & 4 & 1 & 2 & 3 & 4 & 1 & 2 & 3 & 4 & 1 & 2 & 3 & 4 & Average\\
		\midrule
		Finetune& 29.0 & 156.9 & 189.6 & 182.8 & - & 21.2 & 174.5 & 161.5 & - & - & 11.4& 48.0 &  - & - & - & 12.7 &101.2\\
		EWC~\cite{Kirkpatrick2017} & 29.0 & 147.3 & 190.7 & 186.2 & - & 65.9 & 165.4 & 155.9 & - & - & 38.2 &  48.2 & - & - & - & 33.8 & 106.1\\
		MeRGAN-RA~\cite{wu2018memory} & 29.0 & 56.4 & 58.2 & 61.3 & - & 50.9 & 53.7 & 65.9 & - & - & 23.2 & 28.3 & - & - & - & 15.7 &  42.8\\
		\midrule
		EFT-$a_{32}b_{16}$ (+4.8\%) & 29.0 & 29.0 & 29.0 & 29.0 & - & 44.1 & 44.1 & 44.1 & - & - & 32.3 & 32.3 & - & - & - & 27.2 & \textbf{33.1}\\ \midrule
		\bottomrule
	\end{tabular}
	\vspace{-1.5em}
\end{table*}

\subsection{Expansion and Computation Cost}\label{sec:exp_cost}
\vspace{-0.5em}
While expansion-based methods are able to effectively scale to an arbitrary number of tasks, the number of added parameters needed for each task is an important consideration.
At worst, the model growth per task should not exceed the size of the original architecture, as at that point, it is no more efficient than learning a set of independent full models and then ensembling.
We compare our parameter-efficient formulation with other expansion-based methods using the convolutional architecture of \cite{yoon2020scalable}, showing both final total parameter expansion factor and average accuracy at the end of the CIFAR-100/10 task sequence in Table~\ref{tab:capacity}.
For these experiments, we set $a=5$ and $b$ to the full channel depth ($20$ and $50$ for the first and second layers, respectively).
Notably, our proposed method requires the least amount of expansion by a large margin (only $1.04\times$ more parameters than the base network), while also achieving the highest overall accuracy.

While the number of parameters is representative of the amount of memory necessary for the model, memory is not the only computational consideration; the number of floating point operations (FLOPs) is also an important consideration.
While the two are often correlated, they are not necessarily one-to-one: for example, convolutional layer weights take up only $~10\%$ of the total parameters in VGG-16, but they represent $~90\%$ of the total FLOPs. Thus, we also calculate the increase in FLOPs resulting from adding EFTs with various values of $\{a,b\}$ to a ResNet-18 architecture on CIFAR-100/10 in Table~\ref{tab:flops}. 
In terms of wall clock time, we also observe minimal impact form EFTs.
For example, inference time per task (1K samples) requires 1.05 and 0.87 seconds for EFTs and the original model respectively, for task incremental learning (TIL). 

\subsection{Task Inference Regularization}
\vspace{-0.5em}
We proposed max-margin task-specific feature distribution regularization in Section~\ref{sec:task_pred} to improve task prediction. We conduct an ablation study of the loss term $\mathcal{L_M}$ to demonstrate its effectiveness. We use 10\% of the training data as a validation set to tune hyperparameters, finding $\lambda_1=0.05$ and $\Delta=1$ to work well. Our empirical study (Table~\ref{tab:ablation}) shows that the regularization term $\mathcal{L_M}$ boosts task prediction accuracy.  
\vspace{-0.5em}
\subsection{Generative Modeling}
\vspace{-1mm}
In addition to the discriminative models shown above, we also apply our continual learning approach to deep generative models.
As in classification, if a generative adversarial network (GAN)~\cite{goodfellow2014generative} is trained on a sequence of different data distributions, it will experience catastrophic forgetting~\cite{liang2018generative, zhai2019lifelong,cong2020gan,varshney2021efficient}, tending to forget how to produce samples from previous distributions.
We train a StackGAN-v2~\cite{zhang2018stackgan} on a sequence of different image distributions: cats (ImageNet~\cite{Deng2009}), birds (CUB-200~\cite{WelinderEtal2010}), churches and towers (LSUN~\cite{yu15lsun}).

\begin{figure}
	\centering
	\includegraphics[width=0.48\textwidth]{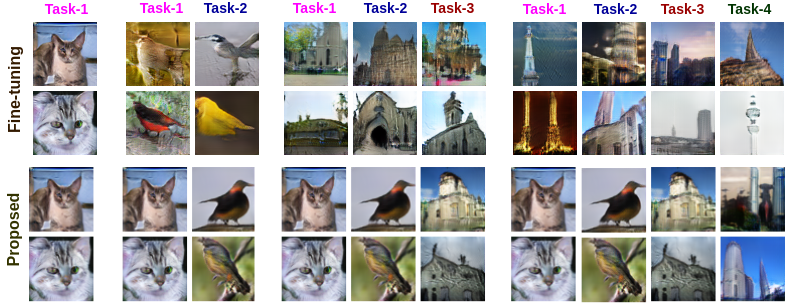}
	\caption{\small Generated samples from the cat, bird, church, and tower distributions after training on each dataset.
		While naive finetuning (top) forgets previously seen distributions, EFT (bottom) effectively retains them.
	}
	\label{fig:gan}
	\vspace{-1.5em}
\end{figure}

After completing training on a dataset, we report Fr\'{e}chet Inception Distance (FID)~\cite{heusel2017gans} to quantify GAN performance for each previously seen dataset (Table~\ref{tab:gan}).
A finetuning approach commits the entirety of the network to learning each new dataset, without any restrictions or regard for previous tasks, and is thus able to attain very low FID scores on each dataset directly after training on them.
However, as expected, we also observe that finetuning directly results in severe catastrophic forgetting, leading to the generator completely forgetting previous data distributions (see Figure~\ref{fig:gan}, top); for the finetune approach, each supercolumn in Table~\ref{tab:gan} shows excellent performance for a task when first trained on, followed by immediate degradation.
Applying the popular regularization approach EWC~\cite{Kirkpatrick2017} does not mitigate this issue; we still see severe forgetting.
On the other hand, we observe that EFT is able to effectively learn multiple data distributions sequentially without forgetting (see Figure~\ref{fig:gan}, bottom), beating MeRGAN-RA~\cite{wu2018memory} in final average FID.

\vspace{-1mm}
\section{Conclusions}
\vspace{-1mm}
We propose Efficient Feature Transforms, balancing the need for expressivity to model incoming tasks with parameter efficiency, allowing the model to expand its knowledge without ballooning its size and computation.
We demonstrate the superiority of EFTs with a thorough slate of varied experiments, including on ImageNet-1K and CIFAR-100, demonstrating performance gains while also being efficient, without having to resort to saving data samples for replay.
We also demonstrate the success on complex generative modeling tasks, demonstrating significant catastrophic forgetting mitigation.
By demonstrating success and scalability in challenging settings, we hope to bring continual learning to more practical applications.

\newpage
{\small
\bibliographystyle{ieee_fullname}
\bibliography{egbib}
}

\clearpage
\appendix
\section{Implementation Details}
We describe here the architectures, hyperparameters, and other implementation details used in our experiments.
\subsection{Class Incremental Learning}
\label{apx:res18_cifar100}
CIFAR-100~\cite{Krizhevsky2009} is a medium-scale dataset widely used for supervised classification. 
It contains 100 diverse classes that can grouped into 20 superclasses. We leverage the CIFAR-100 dataset for the class incremental learning setup. We perform experiments in three settings: CIFAR-100/10 which contains 10 task of 10 classes each, CIFAR-100/20 which contains 20 task of 5 classes each and CIFAR-100/5 which contains 5 task of 20 class each. CIFAR-100/20 tests the model's ability to resist catastrophic forgetting for a large number of task, while CIFAR-100/5 tests the model's ability when each task contains more classes.
Interestingly, we observe that many models from the literature showing good results for a small number of tasks perform badly for a larger number of tasks. Thus, we find a large number of tasks to be a more reliable setting to test a model's continual learning ability.

\begin{figure*}[th]
	\centering
	\includegraphics[scale=0.4]{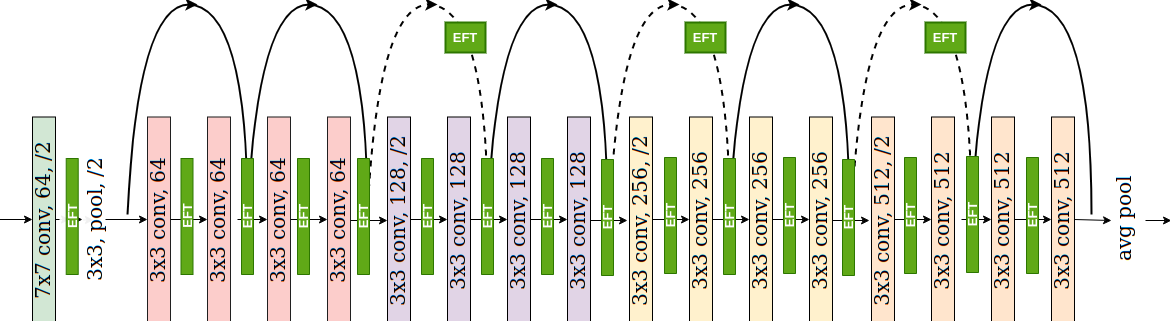}
	\includegraphics[scale=0.45]{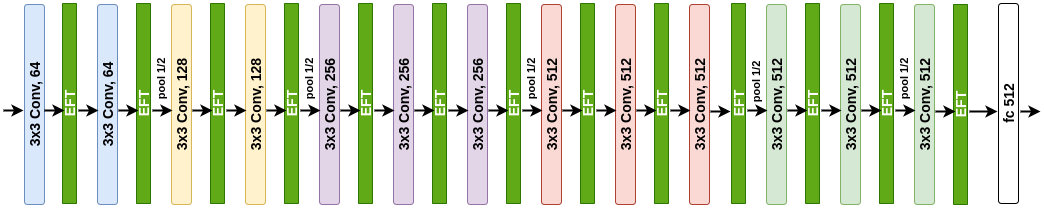}
	\caption{Overview of inserted transformations for ResNet-18. The added EFT layers in each architecture used for continual learning are shown in green. Except for the EFTs, all parameters remain unchanged between tasks.
	}
	\label{fig:eft_arch}
	\vspace{-1em}
\end{figure*}

We use the ResNet-18~\cite{He2016} architecture for all experiments on the CIFAR dataset.
ResNet-18 contains 11.69M parameters across 18 layers. The proposed \emph{Efficient Feature Transformation} (EFT) module can be applied in two ways: $i)$ Serial adaptation and $ii)$ parallel adaptation (see Section~\ref{sec:parallel}). 
The added serial adapter on ResNet-18 architecture is shown in the Figure~\ref{fig:eft_arch} (top).

We use the SGD optimizer for all experiments with a weight decay of $0.005$. The ResNet-18 architecture is trained for 250 epochs for the first task and 200 epochs for subsequent tasks, as forward transfer from the first task leads to quicker convergence. We train the first task with an initial learning rate of $0.01$, with $0.1\times$ learning rate decay at epochs 100, 150, and 200. For the second task onward, the same initial learning and decay rate are used, with decay steps occurring at epochs 70, 120 and 150. In all our experiments (CIFAR-100/10, CIFAR-100/20 and CIFAR-100/5), we set $\lambda=0.05$. For hyperparameter tuning, 10\% data was held out as validation data; for final training, we merge the training data and validation data. In Equation~\ref{eq:add_H}, $\gamma=0$ indicates that the $1\times 1$ filters are not used for the EFT transformation. We show a setting with such a case in Tables~\ref{tab:flops} and \ref{tab:c100_parallel}. In the class incremental learning scenario, task IDs are available during training, but they must be inferred at test time to determine the corresponding $\tau_t$. During training, only samples of task $t$ are present, while at test time, test samples may come from any of tasks $1$ to $t$.

\subsection{Task Incremental Learning}
We perform experiments in the task incremental learning scenario on medium- and large-scale datasets with two standard architectures: VGG-16~\cite{simonyan2014very} and AlexNet~\cite{krizhevsky2012imagenet}. 

\textbf{ImageNet-1K} The ImageNet-1K~\cite{Deng2009} classification dataset contains 1000 classes based on the WordNet~\cite{miller1998wordnet} hierarchy. In the continual learning setup, 1000 classes are divided into 10 tasks of 100 classes each. 
We use the SGD optimizer to optimize an AlexNet architecture. The model is trained for 90 epochs for ImageNet with a weight decay of $0.0005$ for the first task and a weight decay of $0.00005$ for subsequent tasks. An initial learning rate of $0.0001$ is used with a step decay of $0.5\times$ at epochs 40, 60 and 80. EFT layers are appended after each layer.

\textbf{Tiny-ImageNet-200} The Tiny-ImageNet-200 dataset is a subset of the ImageNet dataset and contains 200 classes at downsampled resolution. For continual learning, these 200 classes are divided into 10 tasks with 20 classes each. Since Tiny ImageNet has a lower resolution, the last maxpool layer and the last three convolutional layers from the feature extractor are removed from the standard VGG-16 architecture~\cite{masana2020ternary}. The VGG-16 model is trained for $160$ epochs with an initial learning rate of $0.01$ and a weight decay of $0.0005$, for all tasks. The learning rate is decayed by a factor of $0.1\times$ at epochs 70, 100, and 120. EFT layers are appended as shown in Figure~\ref{fig:eft_arch}.

\subsection{VGG-16 for Heterogeneous Datasets}
Many previous approaches evaluate the model's performance in homogeneous settings: tasks strongly resemble the preceding ones.
We evaluate the model on heterogeneous datasets, where later tasks are very different from previous ones, and the number of classes may change between tasks. 
In this challenging setup, we use the VGG-16 architecture with batchnorm to learn CIFAR-10, SVHN, and CIFAR-100 in sequence. 
We show results for two different task orders: CIFAR-100$\rightarrow$CIFAR-10$\rightarrow$SVHN and SVHN$\rightarrow$CIFAR-10$\rightarrow$CIFAR-100. 
We report the results in Table~\ref{tab:hetero} as the final performance of the model on each of the datasets at the conclusion of the task sequence. The VGG-16 architecture is trained for 200 epochs with an initial learning rate of $0.01$ and a weight decay of $0.0005$. A learning rate decay of $0.1\times$ is used at steps 100, 150, and 170. 

\subsection{Parallel Adaptation}
An alternative parallel adaptation strategy is proposed in Section~\ref{sec:parallel}. We show results for the parallel adaptation strategy in Table~\ref{tab:c100_parallel}. We observe that parallel adaptation with EFTs results in slightly inferior performance compared to serial adaptation. Note that for both parallel and serial adaptation, we use the same optimizer, learning rate, learning rate decay, and weight decay. Please refer to Section~\ref{apx:res18_cifar100} for more details.

\begin{table*}[!th]
	\small
	\centering
	\addtolength{\tabcolsep}{0.9pt}
	\caption{\small Average accuracy on CIFAR-100 in class incremental learning setting when trained on 10 tasks sequentially.}
	\vspace{0.3em}
	\begin{tabular}{l l  l l l l l l l l l l}
		\toprule
		{Dataset / \#Tasks} &{Methods} & {1} & {2} & 3 & 4 & 5 & 6 & 7 & 8 & 9 & Final \\
		\midrule
		\multirow{9}{*}{CIFAR-100/10} & 
		{Finetune} &  88.5& 47.1& 32.1& 24.9& 20.3& 17.5& 15.4& 13.5& 12.5& 11.4\\
		& FixedRep &  88.5& 45.9& 30.1& 22.4& 17.7& 15.2& 12.3& 11.1& 9.8& 8.8\\
		&LwF~\cite{li2017learning} &  88.5& 70.1& 54.8& 45.7& 39.4& 36.3& 31.4& 28.9& 25.5& 23.9\\
		&EWC~\cite{Kirkpatrick2017} &  88.5 & 52.4& 48.6& 38.4& 31.1& 26.4& 21.6& 19.9& 18.8& 16.4\\
		&EWC+SDC~\cite{yu2020semantic} &88.5&78.8&75.8&73.1&71.5&60.7& 53.9&43.5&29.5&19.3\\
		&SI~\cite{Zenke2017}  &  88.5& 52.9& 40.7& 33.6& 31.8& 29.4& 27.5& 25.6& 24.7& 23.3\\
		&MAS~\cite{Aljundi2018} &  88.5& 42.1& 36.4& 35.1& 32.5& 25.7& 21.0& 19.2& 17.7& 15.4\\
		&RWalk~\cite{chaudhry2018riemannian}  &  88.5& 55.1& 40.7& 32.1& 29.2& 25.8& 23.0& 20.7& 19.5&17.9\\
		&DMC~\cite{zhang2020class}  & 88.5& 76.3& 67.5& 62.4& 57.3& 52.7& 48.7& 43.9& 40.1&36.2\\
		\cmidrule{1-12}
		&EFT-$a_4b_{0}$ (+1.7\%) & 90.3&73.9& 65.9& 59.0&54.6&50.6&48.4&45.6&43.3&\textbf{41.2}\\
		&EFT-$a_4b_8$ (+2.0\%) & 90.3&73.9& 65.9& 59.0&54.6&50.6&48.4&45.6&43.3&\textbf{41.5}\\
		&EFT-$a_8b_{16}$ (+3.9\%) & 90.3& 74.2 & 66.5& 60.4&55.8 &51.5& 49.5&46.7& 44.7& \textbf{42.7}\\
		\bottomrule
	\end{tabular}
	\label{tab:c100_parallel}
\end{table*}

\subsection{StackGAN-v2 for Generative Modeling}
Continual learning in generative models (\textit{e.g.}, GANs) is a challenging problem rarely explored by recent literature. 
Existing methods tend to rely on replay-based approaches. 
In contrast, we use an expansion-based method to learn to generate successive datasets in a continual fashion. 
Specifically, we use StackGAN-v2~\cite{zhang2018stackgan} as the base network and use EFT layers to continually expand the architecture for each novel task. 
We append EFT layers after each convolutional layer (serial adaptation) in both the generator and the discriminator, before batchnorm and ReLU activations are applied, resulting in 4.8\% extra parameters per dataset. We use the default StackGAN-v2 hyperparameter values for learning rate and the number of epochs. 

\section{Others Comparisons}
\begin{figure}
    \centering
    \includegraphics[scale=0.38]{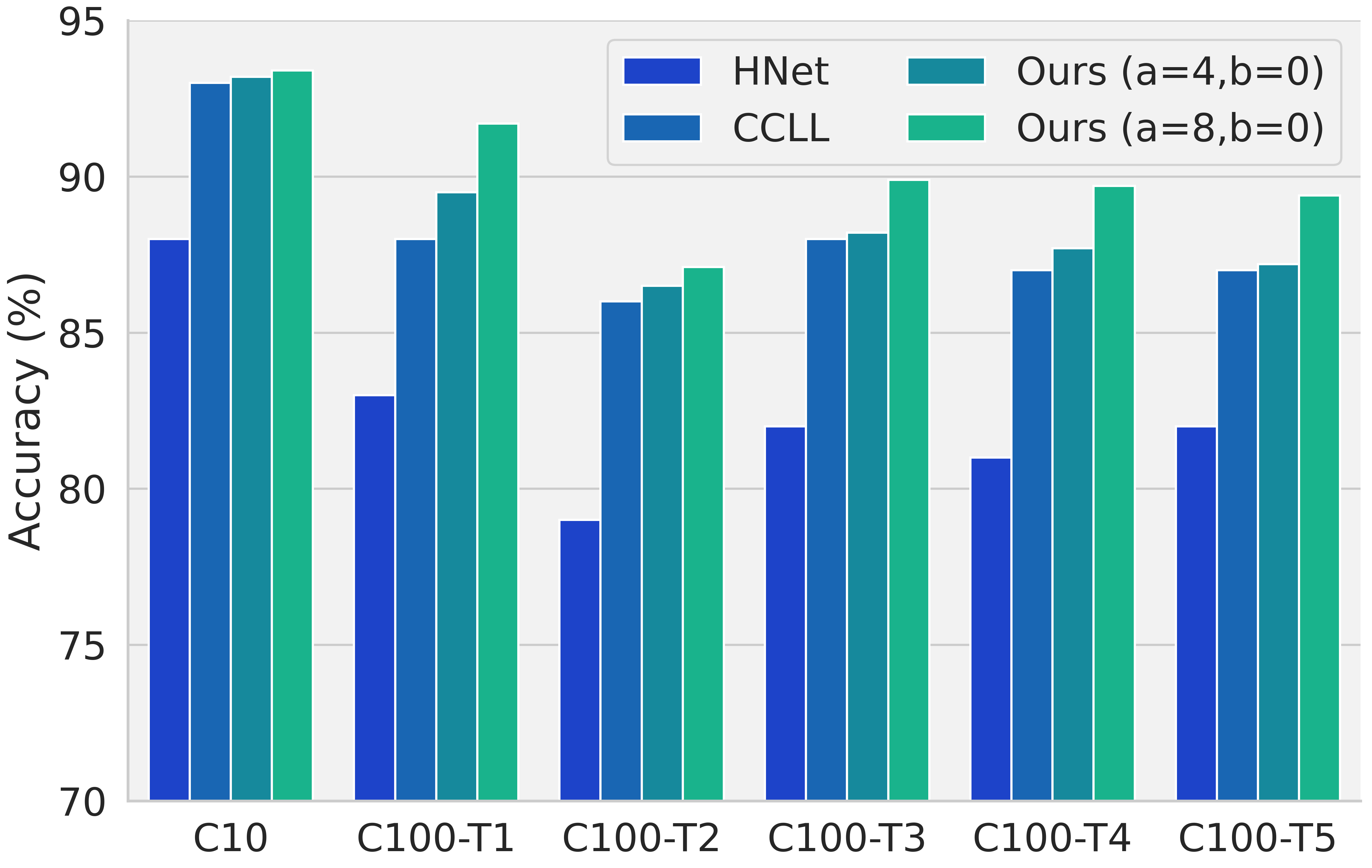}
    \caption{Comparison of the proposed model on the ResNet-32 architecture on the CIFAR10-CIFAR100 dataset.}
    \label{fig:c10-100}
\end{figure}

The model architectures used for continual learning tend to vary from paper to paper, so we did our best to compare our proposed approach with recent baselines in a fair, consistent setting.
However, other papers occasionally report results with other setups.
For example, continual learning performance on CIFAR is occasionally reported with CIFAR-10 being the base dataset, with 5 subsequent tasks of 10 classes drawn from the CIFAR-100 dataset.
For this setting, we follow prior work in using ResNet-32~\cite{He2016}, an architecture with a smaller number of parameters and FLOPs commonly used for CIFAR. 
We compare our proposed approach with the recent works HNet~\cite{von2019continual} and CCLL~\cite{singh2020calibrating} in this setting. The results are shown in Figure~\ref{fig:c10-100}. We use the same training procedure as discussed we did with the ResNet-18 architecture.
We see stronger results with our EFT approach.

ResNet-18/3~\cite{Zenke2017}, a standard ResNet-18 but with a third of the filters per layer, is another widely used architecture for  continual learning. 
We compare our EFTs with the recently proposed SupSup~\cite{wortsman2020supermasks} and BatchE~\cite{wen2020batchensemble} on CIFAR-100/20 in the task incremental setting, showing the results in Table~\ref{tab:supsup}. Again, we observe the proposed model shows significant improvements compared to the baseline model.

\begin{table}[h]
\centering
\addtolength{\tabcolsep}{10pt}
\begin{tabular}{ll}
\toprule
                                          Entry &         Avg Acc@1 \\
\midrule
Upper Bound &  $91.62~\pm 0.89$ \\
SupSup ($\ensuremath{\mathsf{GG}}$)~\cite{wortsman2020supermasks} &  $86.45~\pm 0.61$  \\
SupSup ($\ensuremath{\mathsf{GG}}$) Transfer~\cite{wortsman2020supermasks}  &  $88.52~\pm 0.85$  \\
BatchE ($\ensuremath{\mathsf{GG}}$)~\cite{wen2020batchensemble} &  $79.75~\pm 1.00$  \\
Separate Heads &  $70.60~\pm 1.40$  \\
\midrule
EFT $a_4b_0 (+5.2\%) $&89.25$~\pm$ 0.31\\
EFT $a_5b_0 (+6.7\%)$&90.17$~\pm$ 0.47\\
\bottomrule
\end{tabular}
\vspace{0.3em}
\caption{Task incremental learning accuracy on CIFAR-100/20 with the ResNet-18/3~\cite{Zenke2017} architecture.}
\label{tab:supsup}
\end{table}

\section{Selection of Hyperparameter $a$ and $b$}
The choice of the hyperparameters $a$ and $b$ control both the model's representational power per task, as well as the growth in parameters and FLOPs. This leads to a trade-off: increasing the value of $a$ and $b$ will increase the model's performance at the cost of higher computation and memory (refer to Table-\ref{tab:flops} for ablation). Optimal values depend on the situation and can vary based on architecture. We observe that for ResNet~\cite{He2016} and VGG~\cite{simonyan2014very}, we can use lower values $a$ and $b$, leading to a parameter growth of $2$-$4\%$, while for the AlexNet~\cite{krizhevsky2012imagenet} architecture, higher values of $a$ and $b$ only lead to $0.6\%$ growth, as a significant proportion of the model parameters are in the fully connected layer. Using lower hyperparameter values for AlexNet results in a very small number of parameters per task, making it difficult for the model to adapt to novel tasks. Additionally, the proposed EFT for fully connected layers only adds a diagonal matrix for continual adaption, which adds only a negligible number of parameters. 
We observe that architectures with more parameters in fully connected layers (especially AlexNet) have a harder time adapting features with just diagonal matrices. Therefore, we find it advantageous to increase the number of convolutional layer parameters. We can observe this in Table~\ref{tab:ImageNet_tiny} for the AlexNet architecture.     

\end{document}